# FFINO: Factorized Fourier Improved Neural Operator for Modeling Multiphase Flow in Underground Hydrogen Storage

Tao WANG [a], Hewei TANG [*, a, b]

[a] Department of Petroleum & Geosystems Engineering

The University of Texas at Austin

Austin, Texas 78712-1585, USA

[b] The Oden Institute for Computational Engineering & Sciences

The University of Texas at Austin

Austin, Texas 78712-1229, USA

* hewei.tang@austin.utexas.edu

# ABSTRACT

Underground hydrogen storage (UHS) is a promising energy storage option for the current energy transition to a low-carbon economy. Fast modeling of hydrogen plume migration and pressure field evolution is crucial for UHS field management. In this study, a new neural operator architecture, factorized Fourier improved neural operator or FFINO is proposed as a fast surrogate model for multiphase flow problems in UHS. Experimental relative permeability curves reported in the literature are also parameterized as key uncertainty parameters for the FFINO model. FFINO model performance with the state-of-the-art Fourier-enhanced multiple-input neural operators or FMIONet model are systematically studied through a comprehensive combination of metrics. Our new FFINO model has 38.1% fewer trainable parameters, 17.6% less training time, and 12% less GPU memory cost compared to FMIONet. The FFINO model also achieves a 9.8% accuracy improvement in predicting hydrogen plume in focused areas, and 16.3% higher accuracy in predicting pressure buildup. Sensitivity analysis identifies that the most influential input parameter to models' performance is the injection rate $Q$, while other parameters show moderate to minor impacts. The inference time of the trained FFINO model is 7,850 times faster than a numerical simulator, which guarantees its superior time efficiency. The novel FFINO model can serve as a fast, accurate, and stable alternative to estimate the temporal and spatial evolution of hydrogen plumes and pressure distributions for real-time UHS applications.

## 1. Introduction

The past decade has witnessed the ongoing energy transformation from carbon-intensive fuels to options with smaller carbon footprints [1-2]. Hydrogen, due to its highest energy density by mass and no carbon emission during combustion, has demonstrated great potential as an energy carrier for a decarbonized future [3]. It is crucial to securely store large quantities of hydrogen, and saline aquifers provide abundant spaces for underground hydrogen storage or UHS [4-6]. The storage of hydrogen in saline aquifers involves multiphase flow which is a phenomenon also encountered in other subsurface processes such as contaminant transport, carbon sequestration, hydrocarbon extraction, and nuclear waste disposal [7-10]. Historically, numerical simulators are leveraged to solve the mass and energy conservation equations of multiphase flow processes [11, 12]. However, the adoption of numerical simulations to solve multiphase flow applications can be time-consuming and computationally intensive. A 2D subsurface multiphase flow case may take up to 10 minutes, while a 3D problem in a similar context can cost 2 hours or more, depending on the complexity of the problem, numerical methods, and computing power [13-17]. Given these drawbacks, scenarios where fast decision making or inverse modeling involving thousands of forward simulations demand alternatives for numerical simulators [18-21].

Most recently, deep learning methods such as neural networks and neural operators have been developed to provide faster substitutes for numerical simulations for subsurface multiphase flow, which are generally referred to as surrogate models [13-16, 18, 22-25, 31-35]. Tang et al. (2020) combine a residual U-Net and a convolutional long-short-term memory recurrent network to model oil-water two-phase flow [13]. Wen et al. (2022) propose U-FNO, achieving better performance on gas saturation prediction (MAE: 1.6%, $R^2$ score: 0.981) in a subsurface $CO_2$ storage case than the original FNO (MAE: 2.8%, $R^2$ score: 0.961) and the CNN architecture (MAE: 3.0%, $R^2$ score: 0.955), while on pressure buildup, U-FNO (MAE: 0.68%, $R^2$ score: 0.992) outperforms FNO (MAE: 0.82%, $R^2$ score: 0.989) and CNN (MAE: 0.89%, $R^2$ score: 0.987). However, on the aspect of computational efficiency, CNN model requires only 562 s/epoch, while the time costs for FNO and U-FNO are 711 and 1872 s/epoch respectively [16]. Mao et al. (2025) investigate the performance of U-Net, Fourier neural operators (FNO), and transformers in UHS and discover that U-Net has the highest accuracy ($R^2$ scores: 0.996 and 0.991 for hydrogen saturation and pressure predictions), FNO has the fastest inference speed, reaching nearly 700 references per second [26]. Since 2020, FNO based neural operators have achieved great success as surrogate models in underground multiphase flow problems such as geological carbon storage [14-16]. Despite these achievements, computational inefficiency of these neural operators also emerge over time, including high CPU and GPU memory demands, large trainable parameter size, and slow training speed, especially when dealing with complex input functional spaces consisting of both spatial and scalar parameters [16, 25]. The main reason why FNO and U-FNO struggle to handle complex input parameter spaces is that both models require scalar parameters to have the same shape as spatial parameters, which can be prohibitively demanding for hardware memories if the scalars' dimension is high. To address the issue, Jiang et al. (2024) formulate FMIONet, which integrates MIONet (a DeepONet derivative for multiple inputs via tensor product) and the U-FNO structure. Their work has demonstrated the capability of FMIONet to cut CPU and GPU memories by 85% and 65% during training and inference compared to U-FNO [25]. However, FMIONet is less accurate compared to U-FNO on the same geological carbon storage dataset. The $R^2$ score for FMIONet is 1% lower than that of U-FNO on gas saturation prediction, while for pressure buildup, FMIONet achieves an $R^2$ score of 0.986 while the score for U-FNO is 0.992.

Collectively, these studies confirm great potentials for data-driven deep learning methods to model subsurface multiphase flow, two significant gaps remain. First, Most existing deep learning surrogate models fail to consider relative permeability uncertainty as an input functional space for modeling subsurface multiphase flow processes [13, 16, 22, 24-26]. However, relative permeability functions are crucial in describing subsurface multiphase flow behaviors. Especially for UHS in saline aquifers, the estimation of relative permeability functions is under substantial uncertainties due to limited available experiments [27-30]. Second, from comparison studies of Wen et. al, Jiang et. al, and Mao et. al [16, 25, 26], existing deep learning models exhibit trade-offs between computational efficiency and accuracy. The two limitations encourage us to develop a new neural operator architecture with improved efficiency and accuracy while incorporating relative permeability uncertainty.

In this work, a novel neural operator, FFINO is developed to predict the temporal and spatial evolution of pressure buildup and hydrogen saturation. A comprehensive dataset with various input parameters, including field parameters (permeability, anisotropy, and porosity) and scalar parameters (injection rate and relative permeability coefficients), has been built for training and testing. We demonstrate FFINO's superior performance over FMIONet on prediction accuracy, trainable parameter counts, model size, stability, and speed to train and infer through a comprehensive collection of metrics and paired t-tests. A sensitivity analysis of input parameters is included to quantify the impact of input parameter variations on the models' performance.

The organization of the paper is as follows. In Section 2, the governing equations of UHS multiphase flow, numerical simulation details, and parameter sampling methods are described. In Section 3, the details on the formulation of our novel neural operator and information on model training and evaluation are presented. Subsequently, the results of model performance comparison, visualizations, and sensitivity analysis are presented in Section 4, Section 5 gives the discussions on paired t-tests, limitations, and future work, and Section 6 summarizes the main conclusions of the work.

## 2. Problem setting

### *2.1. Governing equations*

Injecting hydrogen into a subsurface aquifer can be modeled as two-phase flow of gas and brine in porous media. The mass conservation equation for the component $i$ ($i$ can be either hydrogen or brine) is expressed as:

$$\frac{\partial \phi}{\partial t}\left(\sum_j \rho_j x_{ij} S_j\right) + \nabla \cdot \left(\sum_j \rho_j x_{ij} v_j\right) - \sum_j \rho_j x_{ij} q_j = 0, \tag{1}$$

where $x_{ij}$ denotes the mass fraction of the component $i$ in phase $j$, $\rho_j$ is the phase density of the phase $j$, $S_j$ is the saturation of the phase $j$, $\phi$ is the porosity, $q_j$ is the volumetric flux of phase $j$, and the phase transport velocity in porous media $v_j$ is approximated by Darcy's law:

$$v_j = -\frac{kk_{rj}}{u_j}\left(\nabla P_j - \rho_j \mathbf{g}\right), \tag{2}$$

where $P_j$ is the pressure of phase $j$, $k$ is the absolute permeability of the matrix, $k_{rj}$ is the relative permeability of phase $j$, $u_j$ is the phase viscosity of phase $j$, and $\mathbf{g}$ is the gravitational acceleration.

The phase pressure $P_j$ in the system is correlated with capillary pressure $P_c$:

$$P_c = P_n - P_w, \tag{3}$$

where $P_n$ is the pressure of the nonwetting (hydrogen) phase and $P_w$ is the pressure of the wetting (brine) phase.

## *2.2. Numerical simulation*

CMG-GEM (Version 2024.30, Computer Modeling Group Ltd.) is employed to develop multiphase flow datasets for UHS. The hydrogen phase and brine phase are immiscible, and the solubility of hydrogen in brine is negligible [36-38]; thus, hydrogen solubility is not considered. Furthermore, to simplify the models, we assume that there's no water vapor in the gaseous phase. Hydrogen volumetric injection rate is constant for each simulation case, ranging from 25,500 to 255,000 $m^3$/day [4, 6, 39], and the maximum bottom hole pressure (BHP) is 248.21 bar (3,600 psi) into a synthetic radially symmetrical horizontal aquifer $D(r, z)$. The dimension of the aquifer is 97.536 m (320 ft) with 64 equal grids on the vertical or $z$ direction and 30,480 m (100,000 ft) with 192 gradually coarsened grids on the radial or $r$ direction. Fig. 1 shows the grids of the radially symmetrical model. The grid size is optimized to capture the characteristics of hydrogen plume migration and pressure buildup while maintaining computation efficiency. A vertical well with a radius of 7.62 cm (0.25 ft) is perforated at grid (1, 45). The injection period is predetermined to be 180 days [6, 39, 40], divided into 12 reporting time steps on the $1^{st}$, $4^{th}$, $9^{th}$, $16^{th}$, $25^{th}$, $37^{th}$, $52^{nd}$, $70^{th}$, $91^{st}$, $116^{th}$, $145^{th}$, and $180^{th}$ days. The top and bottom boundaries are set as no-flow boundaries, and the outer boundary is set to be closed, and the distance from the injection well to the outer boundary is remote to mimic an infinite-acting aquifer. Meanwhile, the aquifer temperature is set to be 50 °C (122 °F) and the reference depth is set to 914.4 m (3,000 ft), which is also the top of the aquifer. At reference depth, the reference pressure is set to 93.08 bar (1,350 psi). Synthetic simulation results including the temporal and spatial evolution of gas saturation (*sg*) and pressure buildup (*dP*) are used to build the training and testing datasets. The visualizations of results in this study will show all grids in the $z$ direction and grids within the first 1,000 meters on the $r$ direction.

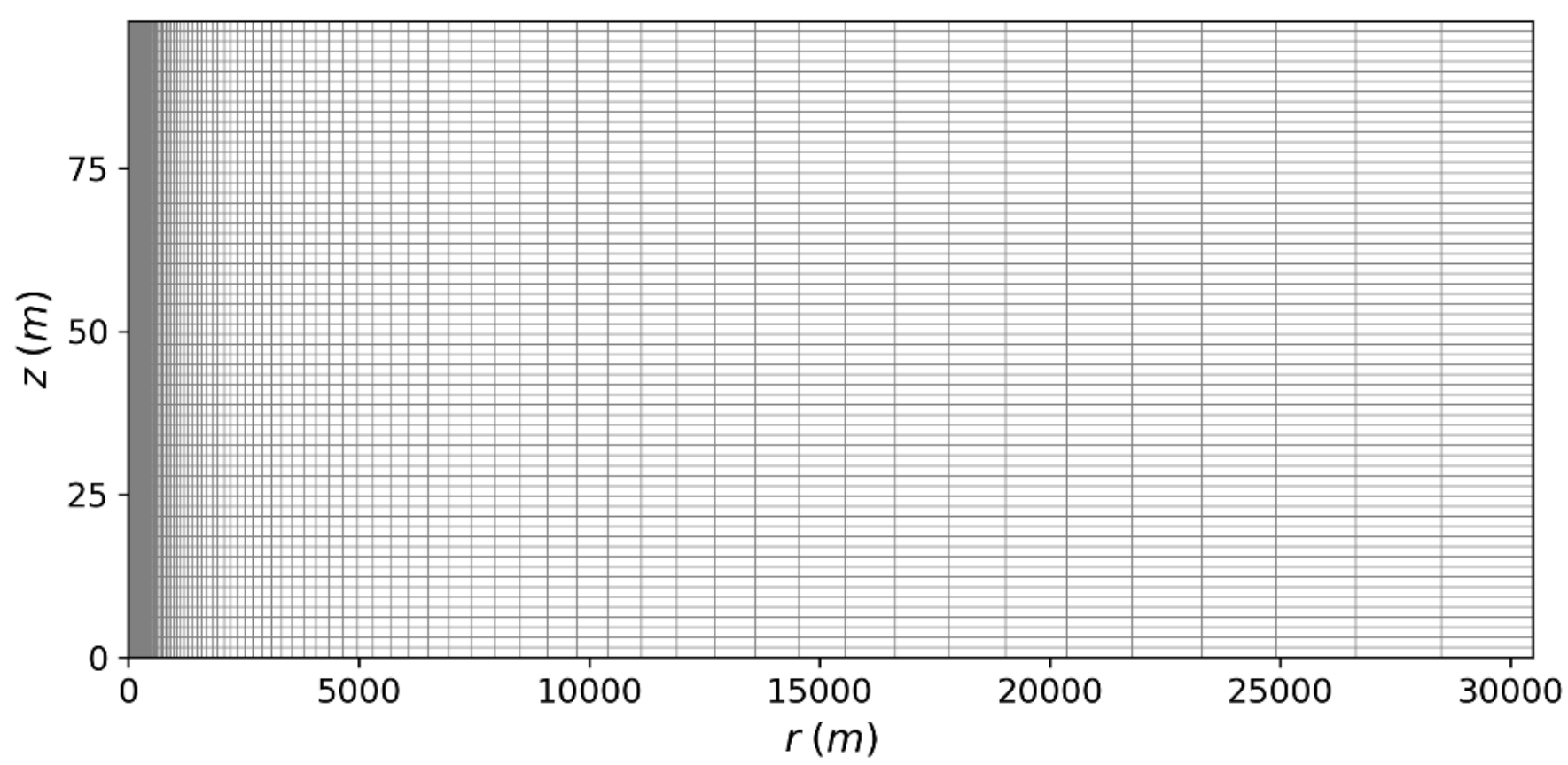

**Fig. 1.** Aquifer model used in this study ($r$: radial distance, $z$: vertical position)

## *2.3 Parameterization of relative permeability*

For UHS, the relative permeability curves are crucial parameters to regulate the behaviors of hydrogen-brine two-phase flow [41-43]. Our current investigation targets the prefill/injection stage for UHS, which is a typical drainage process, and relative permeability hysteresis is thus not considered [28, 29, 44]. The modified Brooks-Corey (MBC) correlations proposed by Lake et al. (1989) are leveraged to fit available hydrogen water/brine drainage relative permeability data [45], which are obtained under similar temperature (18 to 45 °C) or pressure conditions (55 to 100 bar) as our simulation settings. The correlations take the form of the following equations:

$$k_{rw} = k_{rw,max}\left[\frac{S_w - S_{wi}}{1 - S_{gr} - S_{wi}}\right]^m, \quad (4)$$

$$k_{rg} = k_{rg,max}\left[\frac{1 - S_w - S_{gr}}{1 - S_{gr} - S_{wi}}\right]^n, \quad (5)$$

where $S_w$ is the water saturation, $k_{rw}$ is the relative permeability of water at $S_w$, $k_{rg}$ is the relative permeability of water at $S_w$, $k_{rw,max}$ is the end-point relative permeability of the water phase, $k_{rg,max}$ is the end-point relative permeability of the gas phase, $S_{wi}$ is the irreducible water saturation, $S_{gr}$ is the residual gas saturation, $m$ is the water relative permeability exponent, and $n$ is the gas relative permeability exponent. The ranges of modified Brooks-Corey correlation coefficients ($k_{rw,max}$, $k_{rg,max}$, $S_{wi}$, $S_{gr}$, $m$, and $n$) are crucial parameters to determine the shape of hydrogen/brine relative permeability curves, and thus they are key parameters to control underground hydrogen/brine two-phase flow. A least squares algorithm is applied to correlate available hydrogen water/brine relative permeability curves as presented in Fig. 2. A Latin Hypercube Sampling (LHS) algorithm is implemented to sample the coefficient spaces to reconstruct the relative permeability curves for simulations. The LHS algorithm has been widely adopted since its initial release [46-48], especially in high-dimensional parameter spaces. As dimensionality increases, simple random sampling will leave rapidly growing unsampled regions. LHS algorithm mitigates this challenge by splitting each parameter space uniformly and sampling each parameter space independently, resulting in better coverage and more reliable statistical estimates with fewer samples.

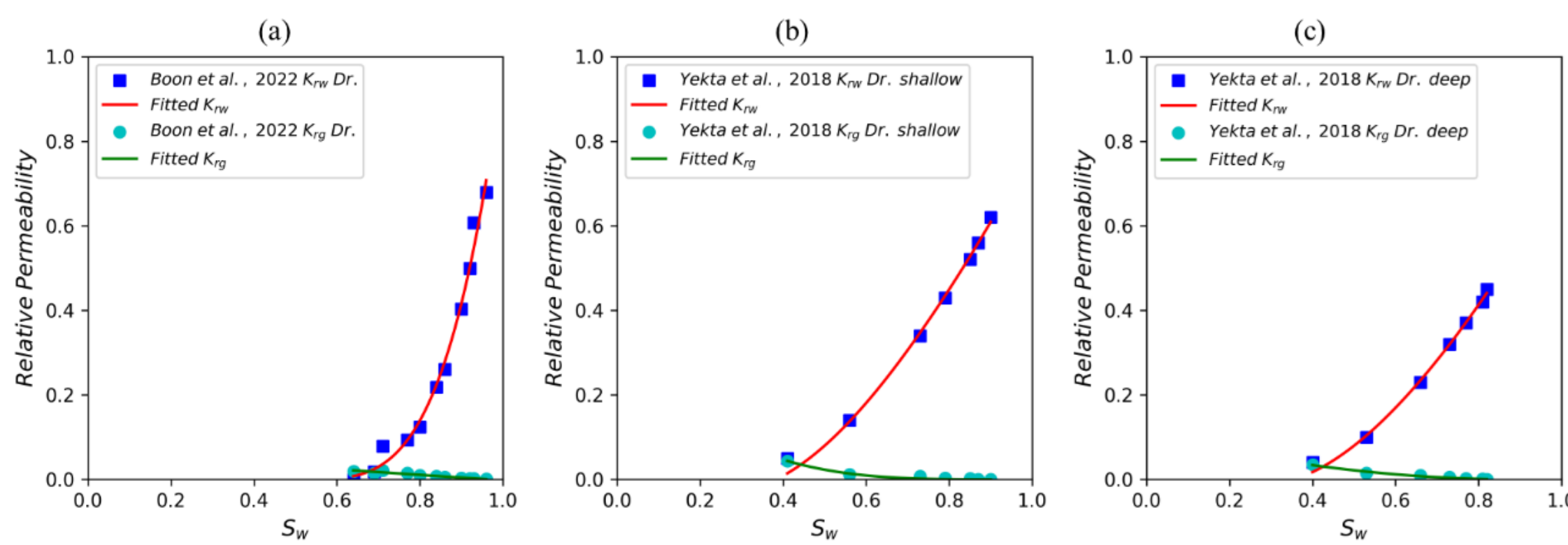

**Fig. 2.** Modified Brooks-Corey correlations of hydrogen water/brine drainage relative permeability used in this study (a: Boon et al. 2022 [28], b: Yekta et al. 2018 shallow case [27], and c: Yekta et al 2018 deep case [27])

### *2.4. Parameter sampling*

The input parameters for the UHS simulation can be categorized into two groups: spatial parameters that vary with changing locations, and scalar parameters that remain constant regardless of locations. Spatial parameters include horizontal permeability ($k_h$), vertical anisotropy ($k_v/k_h$), and porosity ($\phi$).

$k_h$: A fractal algorithm is implemented to generate random heterogeneous $k_h$ maps, previously implemented by Tang et al. [14].

$k_v/k_h$ : The vertical anisotropy map defines the proportion of vertical permeability to that of horizontal permeability. The range of anisotropy is 0.01 to 1.00.

$\phi$: To account for the loose correlation between $\phi$ and $k_h$ [49], a correlation of porosity and permeability previously adopted in Ref. [14] is employed and $\phi$ is perturbed with a random Gaussian noise $\mathcal{N}(0,\ 0.005)$.

Scalar parameters include the injection rate $Q$ and the hydrogen water/brine relative permeability. Fig. 3 gives a schematic illustration of the parameters sampling process.

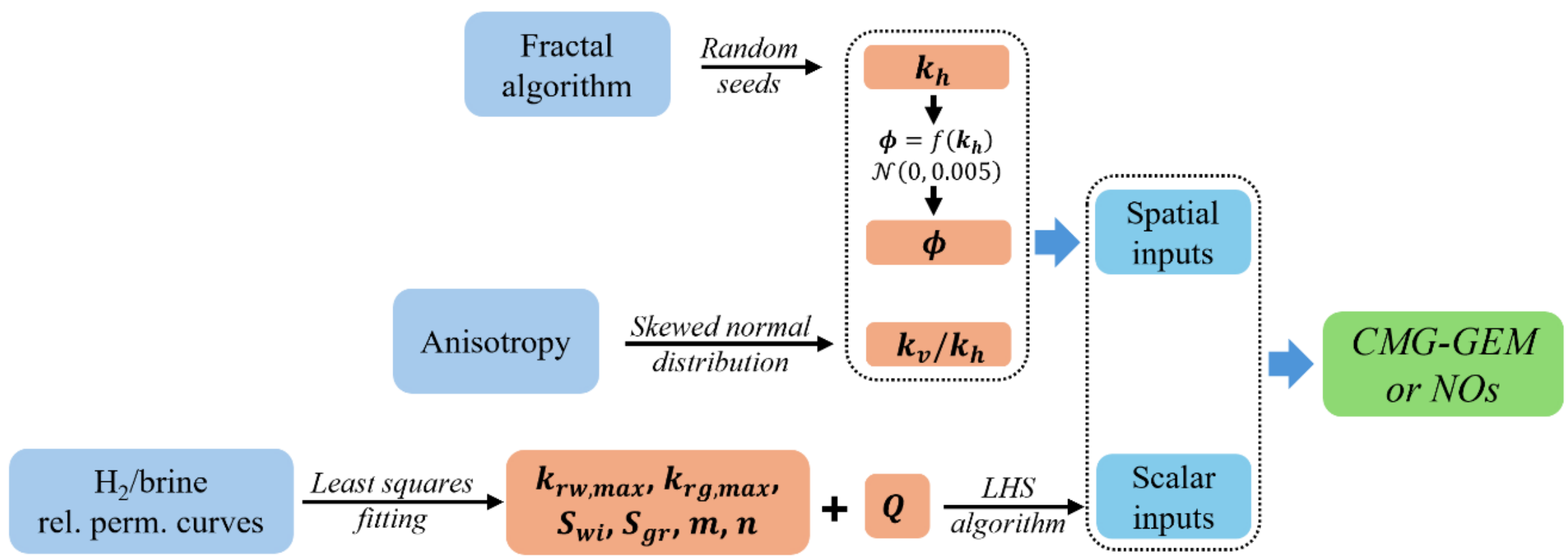

**Fig. 3.** Schematic illustration of the parameters sampling process for CMG simulation and Neural Operators training (*NOs* denotes Neural Operators)

To demonstrate how uncertainties in relative permeability functions influence the simulation results of UHS, the relative permeability curves presented in Fig. 2 are used, while keeping all other spatial and scalar inputs identical. Table 1 summarizes the scalar inputs used for the relative permeability comparison study. The simulation results are shown in Fig. 4 and Fig. 5.

**Table 1.** Scalar inputs used for the comparison study of relative permeability

| example | $Q$ (m$^3$/day) | $k_{rw,max}$ | $k_{rg,max}$ | $S_{wi}$ | $S_{gr}$ | $m$ | $n$ |
|---|---|---|---|---|---|---|---|
| 1 | 169,900 | 0.768 | 0.031 | 0.50 | 0.03 | 3.808 | 1.052 |
| 2 | 169,900 | 0.642 | 0.056 | 0.37 | 0.08 | 1.453 | 3.317 |
| 3 | 169,900 | 0.530 | 0.042 | 0.34 | 0.12 | 1.560 | 1.930 |

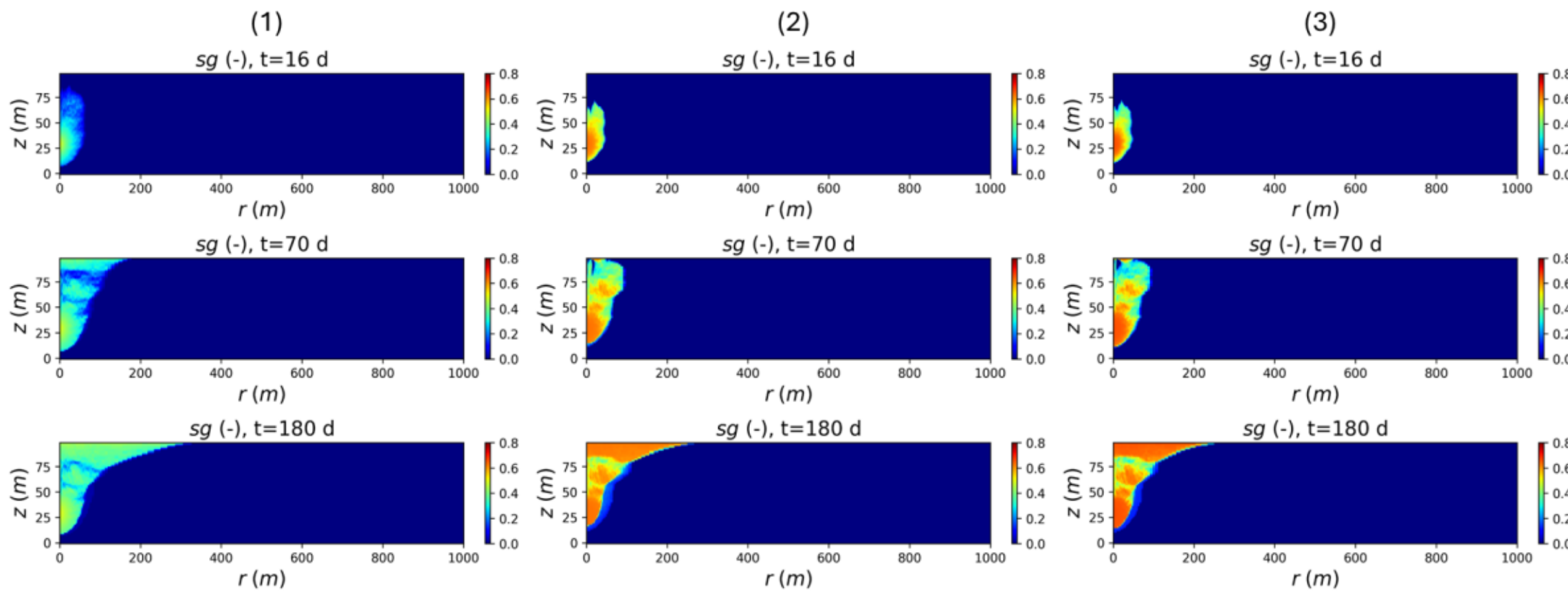

**Fig. 4.** Hydrogen saturation evolution with different relative permeability curves

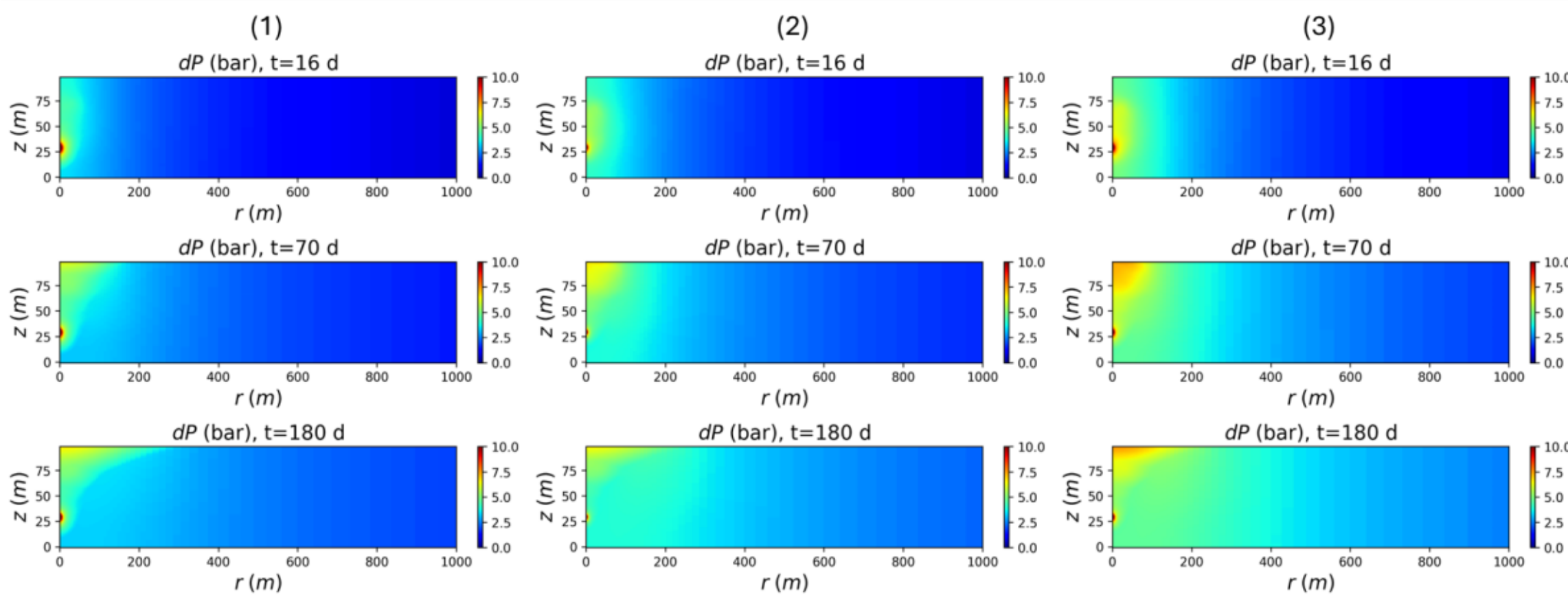

**Fig. 5.** Pressure buildup evolution with different relative permeability curves

Although all three examples have the same inputs except relative permeability coefficients, the hydrogen plume and pressure buildup responses are significantly different. Fig. 4 presents hydrogen plume migration under different relative permeability coefficients. In example 1, a high $S_{wi}$ value limits the hydrogen saturation of the plume, while in examples 2 and 3, a lower $S_{wi}$ enhances hydrogen plume saturation in corresponding grid cells. Hydrogen is simulated to be injected at a constant rate in all three examples, the relatively low hydrogen plume saturation will lead to more extended plume volumes due to mass balance. In all examples, the lower part of plume will shrink after 3 months of injection. It is speculated that the strong buoyance effect of hydrogen will increase the gas saturation in vertical grids as it moves up, thus increasing the vertical relative gas permeability. Then, at one point, the upward hydrogen flow becomes dominant, the lower part of the plume starts to shrink. As shown in Fig. 5, pressure buildup extends beyond the size of the hydrogen plume. The relative permeability coefficients in examples 2 and 3 allow pressure buildup to transmit to regions remote from the extent of the hydrogen plume, while in example 1, pressure buildup is less extensive, and the pressure buildup region almost matches the shape of the plume. With different relative permeability coefficients, hydrogen saturation distribution, plume shape, and pressure buildup are all significantly altered, which supports our proposal to include relative permeability uncertainties in our study.

**Table 2**. Summary of the sampled spatial and scalar parameters used in this study

| Type | Variable | Notation | Range | Unit |
|---|---|---|---|---|
| spatial | Horizontal permeability | $k_h$ | [44.1, 1,000.0] | mD |
| | Anisotropy | $k_v/k_h$ | [0.01, 1.00] | – |
| | Porosity | $\phi$ | [0.140, 0.345] | – |
| scalar | Injection rate | $Q$ | [25,500, 255,000] | $m^3$/day |
| | Water end-point relative permeability | $k_{rw,max}$ | [0.530, 0.768] | – |
| | Gas end-point relative permeability | $k_{rg,max}$ | [0.031, 0.056] | – |
| | Irreducible water saturation | $S_{wi}$ | [0.340, 0.500] | – |
| | Residual gas saturation | $S_{gr}$ | [0.030, 0.120] | – |
| | Water relative permeability exponent | $m$ | [1.453, 3.808] | – |
| | Gas relative permeability exponent | $n$ | [1.052, 3.317] | – |

The type of spatial and scalar parameters, sampling ranges, distributions, dimensions, and units are summarized in Table 2. A more detailed statistical description of spatial and scalar parameters used in the study is given in Appendix A. The spatial inputs have a resolution of (192, 64), while the hydrogen saturation and pressure buildup outputs have a resolution of (192, 64, 12), where 12 represents the reporting time steps. The dataset used for model training contains 3,200 input and output pairs, with the first 3,000 pairs used for model training and the last 200 pairs used for testing. In Fig. 6, an illustration of the inputs, model structure, and outputs of the study is given.

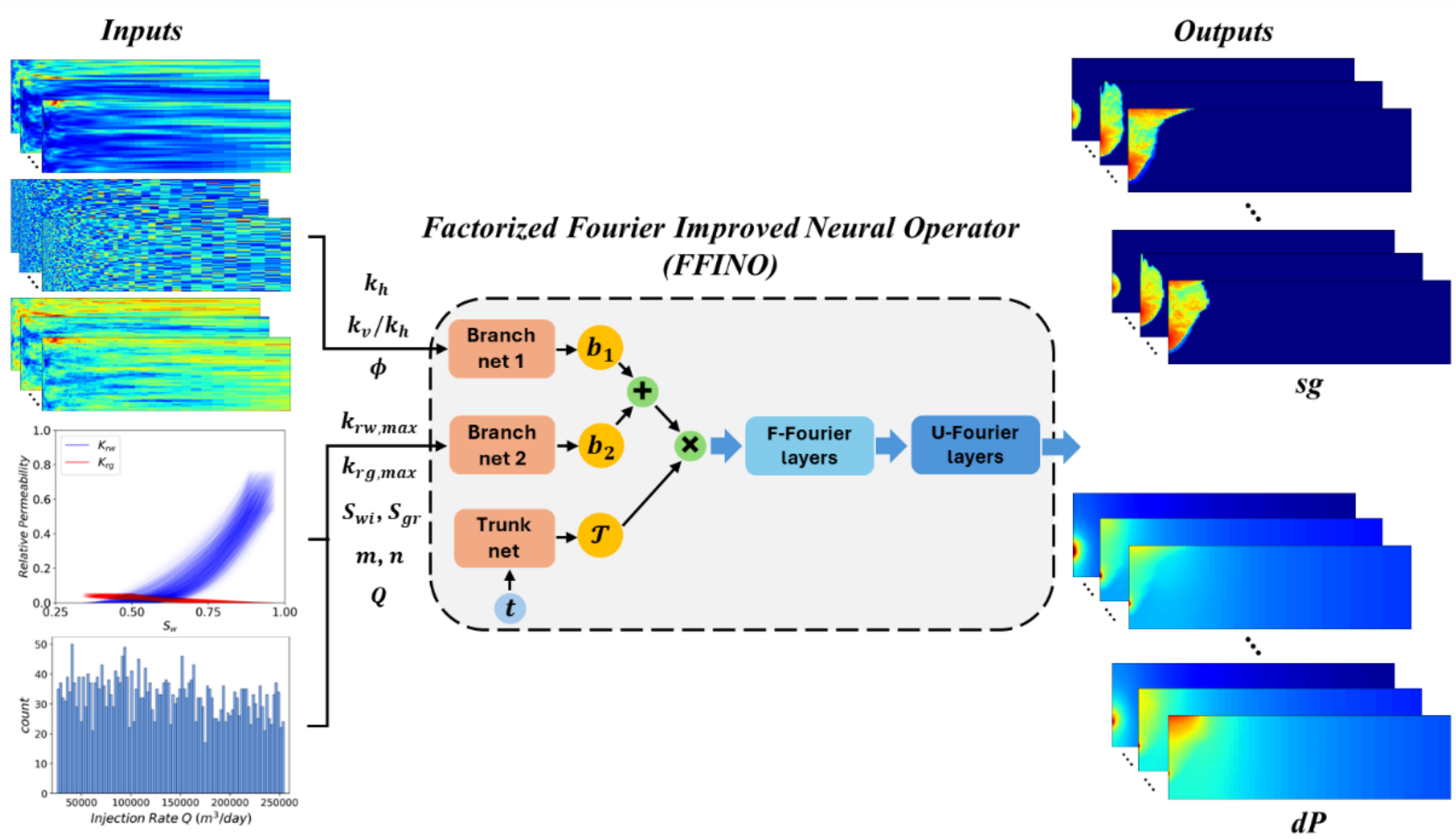

**Fig. 6.** An example of inputs and outputs used for model training includes spatial and scalar inputs, gas saturation (*sg*) outputs, and pressure buildup (*dP*) outputs

## 3. Methods

Neural operators are a group of deep learning architectures that learn infinite-dimensional functional space mappings based on a finite collection of input and output pairs [16, 25, 31-35]. Solving multiphase flow problems of UHS defined on domain $D \subset \mathbb{R}^d$ means finding a nonlinear mapping $\mathcal{G}^*$ from $\mathcal{A}$ (input function spaces $\subset \mathbb{R}^{d_a}$) to $\mathcal{U}$ (output function spaces $\subset \mathbb{R}^{d_u}$) that satisfies corresponding governing equations. Then a neural operator $\mathcal{G}_\theta$ will learn a mapping of $\mathcal{G}^*$ using $\boldsymbol{a}(x) \in \mathcal{A}$ and $\boldsymbol{u}(x) = \mathcal{G}^*(\boldsymbol{a}(x)) \in \mathcal{U}$, where $x$ is the spatial discretization on domain $D$. The approximation of $\mathcal{G}_\theta$ to $\mathcal{G}^*$ is achieved by minimizing a predefined loss function.

### *3.1.MIONet*

MIONet is proposed by Jin et al. (2022) for learning nonlinear operator mapping between functional spaces by leveraging the universal approximation theorem [50]. The operator maps the input functions to the output function as follows:

$$\mathcal{G}: (\boldsymbol{a}_1, \cdots, \boldsymbol{a}_n) \mapsto \boldsymbol{z}, \quad (6)$$

MIONet incorporates n branch nets and one trunk net into its structure. Namely, the $i^{th}$ branch net encodes the corresponding input function $\boldsymbol{a}_i$, while the trunk net encodes the coordinates input $\xi$. Then the mathematical representation of MIONet can be expressed:

$$\mathcal{G}(\boldsymbol{a}_1, \cdots, \boldsymbol{a}_n)(\xi) = \sum_{j=1}^{p} b_j^1(\boldsymbol{a}_1) \times b_j^2(\boldsymbol{a}_2) \cdots \times b_j^n(\boldsymbol{a}_n) \times t_j(\xi) + b_0, \quad (7)$$

where $b_0 \in \mathbb{R}$ is the bias term, $\{b_1^i, b_2^i, \cdots, b_p^i\}$ are the $p$ outputs of the $i^{th}$ branch net, and $\{t_1, t_2, \cdots, t_p\}$ are the $p$ outputs of the trunk net. An illustration of the MIONet structure used in this study is shown in Fig. 7.

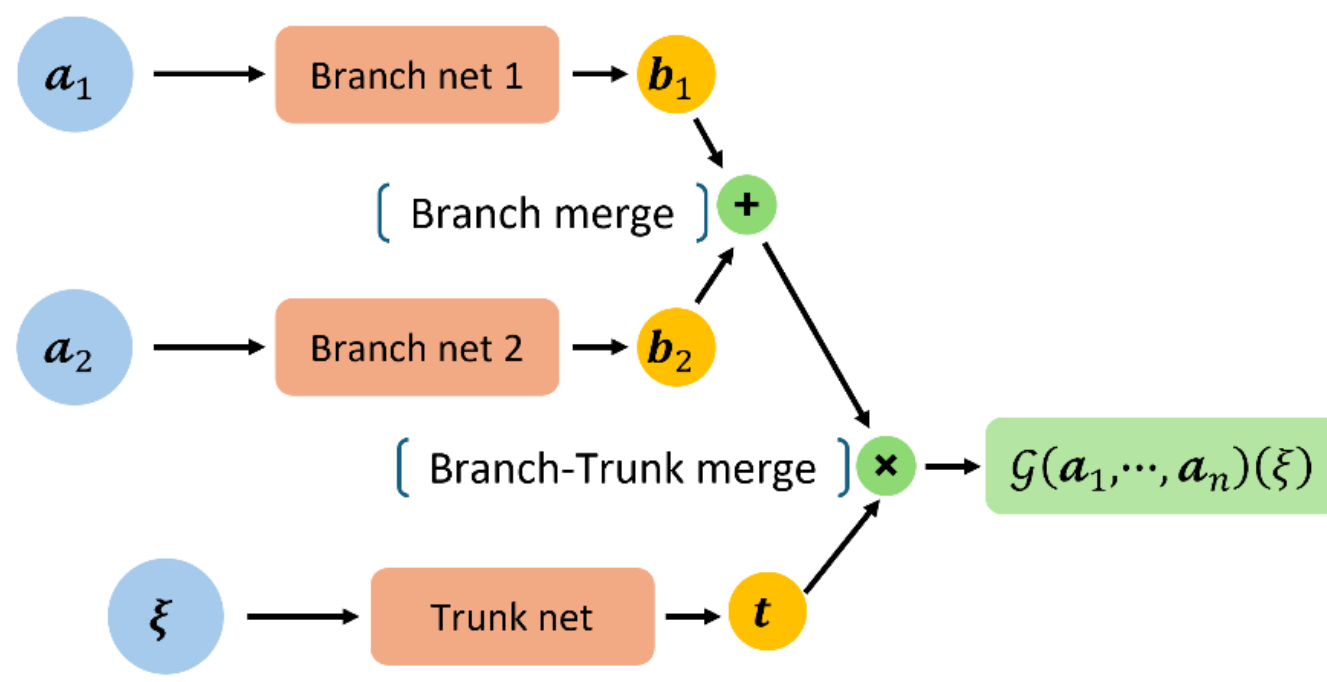

**Fig. 7**. Illustration of the MIONet structure used in this study

### *3.2. Factorized Fourier neural operator (F-FNO)*

Factorized Fourier Neural Operator (F-FNO) is developed by Tran et al. (2021) for learning nonlinear operators mapping between function spaces by separable spectral layers [35]. The input $\boldsymbol{z}$ is passed through a series of operator layers $\mathcal{L}$ to produce the output $\boldsymbol{u}$. We denote each factorized Fourier layer as an F-Fourier layer, and a schematic illustration of an F-Fourier layer structure is shown in Fig. 8a. The overall expression of F-FNO takes the following form:

$$\boldsymbol{u} = \mathcal{G}(\boldsymbol{z}) = \left(\mathcal{Q} \circ \mathcal{L}^{(N)} \circ \cdots \circ \mathcal{L}^{(n)} \circ \cdots \circ \mathcal{L}^{(1)} \circ \mathcal{P}\right)(\boldsymbol{z}), \quad (8)$$

and each F-Fourier layer $\mathcal{L}^{(n)}$ (the $n^{th}$ layer) can be expressed as:

$$\mathcal{L}^{(n)}\left(z^{(n)}\right) = z^{(n)} + \sigma\left[W_2^{(n)}\sigma\left(W_1^{(n)}K^{(n)}\left(z^{(n)}\right) + b_1^{(n)}\right) + b_2^{(n)}\right], \quad (9)$$

### *3.3. U-FNO*

U-FNO integrates additional U-Net blocks in certain Fourier layers in the conventional FNO structure to enhance model accuracy. A block U-Fourier layer can be respectively expressed as:

$$\boldsymbol{z_{m+1}} = \sigma(K_m(\boldsymbol{z_m}) + U_m(\boldsymbol{z_m}) + W_m \cdot \boldsymbol{z_m} + b_{\boldsymbol{m}}), \quad (10)$$

where $K_m$ is the kernel integral linear transformation, $U_m$ is a U-Net operator, and $W_m$ is a linear operator. Fig. 8b shows the general structure of a U-Fourier layer adapted from Ref. [16].

### *3.4. Factorized Fourier improved neural operator (FFINO)*

We formulate a new representation of a neural operator combining MIONet, F-Fourier layers, and U-Fourier layers, denoted as FFINO. An architecture diagram of FFINO is given in Fig. 8. Our major contribution in building the new FFINO structure is that we replace the original Fourier layers in FMIONet with F-Fourier layers. Unlike a Fourier layer on a 2D grid, an F-Fourier layer processes a 2D grid by applying a 1D Fourier transform twice on two factorized dimensions/directions.

The coordinates of the output function $\boldsymbol{u}$ are $x$ and $t$. $x$ is spatial discretization, and $t$ is time discretization. Here, the spatial discretization $x$ is encoded with other spatial inputs in $\boldsymbol{a}_1$, scalar inputs are encoded in $\boldsymbol{a}_2$, and the trunk net encodes the time discretization. To merge results from different branch nets and the trunk net, MIONet employs point-wise summation for the branch merger operation and point-wise multiplication for the branch-trunk merger operation. The point-wise addition and point-wise multiplication are more efficient in handling high-dimensional inputs compared to the direct concatenation operation of multiple functional spaces as adopted in FNO or U-FNO. The overall merger operations can be expressed as:

$$\boldsymbol{z} = (\boldsymbol{b}_1 + \boldsymbol{b}_2)\odot\boldsymbol{\mathcal{T}}, \tag{11}$$

where $\boldsymbol{z}$ is the output for the MIONet structure, $\boldsymbol{b}_1$ is the output of the first branch net, $\boldsymbol{b}_2$ is the output of the second branch net, and $\boldsymbol{\mathcal{T}}$ is the output of the trunk net. Next, a combination of 3 F-Fourier layers and 3 U-Fourier layers are used as a decoder to map from the hidden vector $\boldsymbol{z}$ to the output $\boldsymbol{u}$. Subsequently, a fully connected neural network $Q$ is applied to project the output of the last U-Fourier layer to the output.

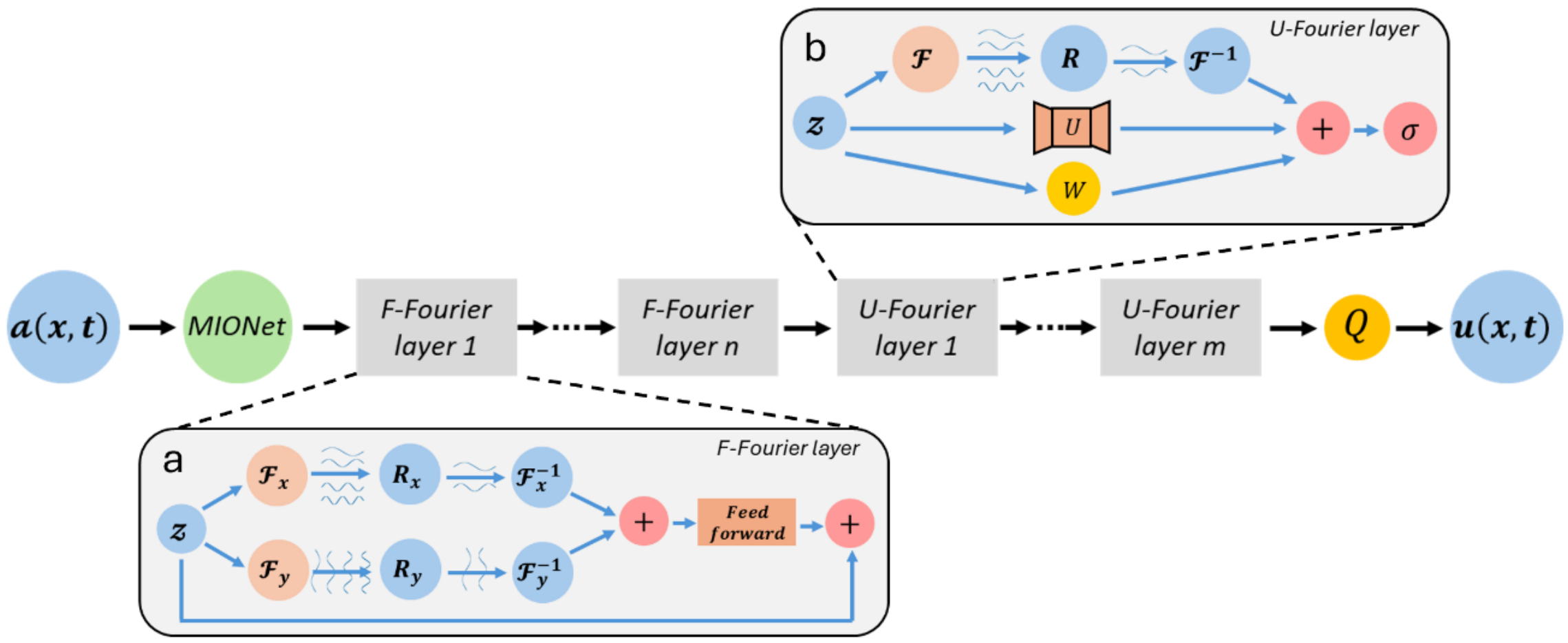

**Fig. 8**. Illustration of the FFINO structure (a: factorized-Fourier layer structure, b: U-Fourier layer structure, n: number of F-Fourier layers, and m: number of U-Fourier layers)

### *3.5. Training and evaluation*

To make a fair comparison with FMIONet, the same $lp$-loss function as in Ref. [16, 25] is leveraged, which is a combination of the relative loss and the first derivative relative loss on $r$ direction and takes the following form:

$$L(y,\ \hat{y}) = \frac{||y-\hat{y}||_p}{||y||_p} + \beta \frac{||\frac{dy}{dr}-\frac{d\hat{y}}{dr}||_p}{||\frac{dy}{dr}||_p}, \tag{12}$$

where $y$ is the reference from numerical simulations, $\hat{y}$ is the prediction from the neural operators, $p$ is the order of the norm set as 2, and $\beta$ is a hyperparameter set as 0.5. The learning rate is initially set as 0.001 and declines at a constant rate.

To evaluate the performance of the trained neural operators, a combination of metrics, including the average $R^2$ score of 200 test samples, average root mean square error (RMSE) of the 200 test samples, mean relative error (MRE), and structural similarity (SSIM) [51] are adopted. Note that MRE and RMSE are evaluated on areas of interest (AOI) where hydrogen saturation is no less than 0.01 and pressure buildup is no less than 0.005 bar. The combination of metrics will help to objectively compare the models' performance on the whole domain and focused areas.

## 4. Results

In this section, the accuracy, stability, and training efficiency of FFINO are demonstrated by comparing with FMIONet, the state-of-the-art model in the literature.

The architecture of the FFINO model is presented in Table 3, and the structure of FMIONet is given in Ref. [25].

**Table 3**. FFINO model structure. "Linear" denotes a linear transformation. "FNN" denotes a fully connected neural network. "Fourier1d/2d" denotes the 1D/2D Fourier transform. "Conv1d" denotes 1D convolution. "UNet2d" denotes a 2D U-Net. In this model, there are 2,282,981 trainable parameters.

| | Layer | Operation | Output shape |
|---|---|---|---|
| Branch net | Branch net 1 | Linear | (4, 64, 192, 36) |
| | Branch net 2 | FNN | (4, 36) |
| Branch merger | – | Point-wise summation | (4, 64, 192, 36) |
| Trunk net | – | FNN | (4, 36) |
| Branch-Trunk merger | – | Point-wise multiplication | (16, 36, 192, 64) |
| Merger net | Projection 1 | Linear | (16, 192, 64, 36) |
| | F-Fourier 1 | Fourier1d/Fourier1d /Add/ReLU/Add | (16, 192, 64, 36) |
| | F-Fourier 2 | Fourier1d/Fourier1d /Add/ReLU/Add | (16, 192, 64, 36) |
| | F-Fourier 3 | Fourier1d/Fourier1d /Add/ReLU/Add | (16, 192, 64, 36) |
| | U-Fourier 1 | Fourier2d/Conv1d/UNet2d/Add/ReLU | (16, 192, 64, 36) |
| | U-Fourier 2 | Fourier2d/Conv1d/UNet2d/Add/ReLU | (16, 192, 64, 36) |
| | U-Fourier 3 | Fourier2d/Conv1d/UNet2d/Add/ReLU | (16, 192, 64, 36) |
| | Projection 2 | Linear | (16, 192, 64, 128) |
| | Projection 3 | Linear | (16, 192, 64, 1) |
| | Reshape | – | (4, 4, 192, 64) |

The structure of MIONet allows the time batch size to be chosen from 1 to the maximum time steps (12 as in this study). Jiang et al. (2024) have done a thorough study to investigate the selection of time batch sizes on the performance of FMIONet [25]. Here, in our study, both sample batch size and time batch size are set as 4 in all experiments. Each model is trained 5 times with the same training dataset for 200 epochs, and the model performance is evaluated using selected comprehensive metrics.

## *4.1. Gas saturation*

FMIONet and FFINO are employed to learn the mapping from input functional spaces to hydrogen plume migration with time. Performance results of the two models on hydrogen saturation are presented in Table 4. The prediction accuracy on hydrogen plumes from FFINO is better across the metrics used in the study. The metrics show that the proposed FFINO has better prediction accuracy and model stability. FFINO is about 10% more accurate than FMIONet based on the RMSE and MRE metrics. FFINO shows significantly smaller standard deviations on all testing metrics from five random training experiments compared to FMIONet, which demonstrates the new model has better model stability.

**Table 4**. FFINO and FMIONet model performance on hydrogen saturation

| *Models* | $R^2$ | *RMSE* | *SSIM* | *MRE* |
|---|---|---|---|---|
| *FMIONet* | 0.9883±0.0009 | 3.44e-2±1.42e-3 | 9.60e-1±3.42e-3 | 9.47e-2±4.42e-3 |
| *FFINO* | 0.9903±0.0003 | **3.12e-2±4.69e-4** | 9.66e-1±7.24e-4 | **8.54e-2±1.13e-3** |

Next, the results of the best-trained FMIONet and FFINO models on hydrogen saturation are elaborated. $R^2$ and MRE distributions across 200 cases in the hydrogen saturation test dataset for the best-trained models are shown in Fig. 9. The $R^2$ and MRE distributions demonstrate that FFINO achieves higher accuracy in most of the 200 cases in the hydrogen saturation test dataset compared to FMIONet. There is one outlier on Fig. 9, on which both FMIONet and FFINO have the worst performance among the 200 cases in the test dataset on $R^2$ and MRE. After analyzing the simulation results, it reveals that the hydrogen plume moves rapidly up near the wellbore while the plume base shrinks dramatically, resulting in relatively larger proportion of saturation fronts compared to the total plume size as shown in Fig. 10, case 1. Current neural operators are struggling to make very accurate predictions on these fronts/edges as revealed by previous studies [16, 25]. Thus, the large ratios of front areas to total plume areas contribute to increased model prediction errors. Because the FMIONet model has more trainable parameters and a larger model size, it can handle this rare and extreme case slightly better.

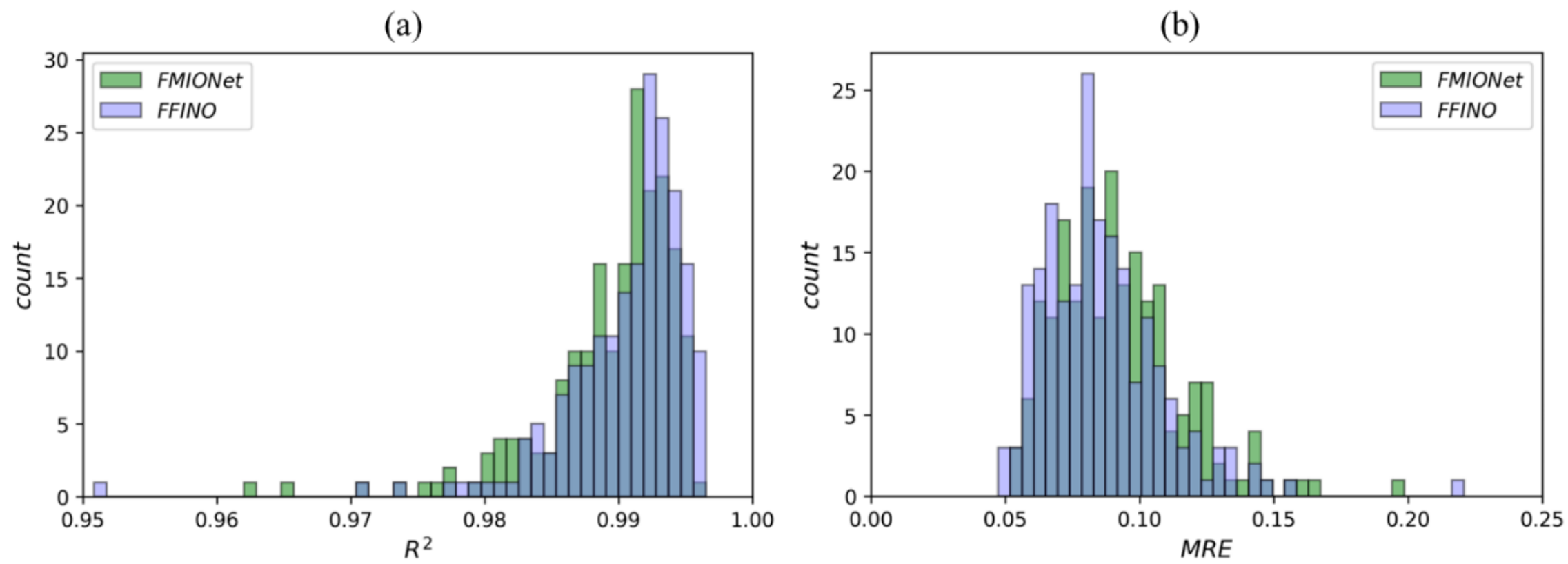

**Fig. 9**. $R^2$ and MRE distributions of 200 cases in hydrogen saturation testing for the best-trained FMIONet and FFINO models (a: $R^2$, b: MRE)

Apart from comparing the average performance of the two best trained models on all hydrogen saturation test samples, visualizations on 2 test cases with different characteristics and complexity are presented in Fig. 10. Each example shown in Fig. 10 gives a comparison of reference hydrogen saturation from numerical simulation, predictions from best trained FMIONet and FFINO models, and corresponding model errors (model predictions subtract simulation references, same for pressure buildup) on each grid cell at the selected time step. During our inspection of the models' predictions, it is discovered that models' performance may vary with time steps. The plume fronts are changing with time, resulting in different ratios of front areas to total plume areas. As these plume fronts account for the major source of model prediction errors, this phenomenon leads to models' performance variation across time steps. In case 1, both models show lower prediction accuracy on the 180$^{th}$ day, the corresponding $R^2$ and MRE values for FMIONet are 0.937 and 0.229, while the two metrics for FFINO are 0.907 and 0.281. In case 2, both models perform well and FFINO is slightly more accurate. FMIONet and FFINO obtain $R^2$ scores of 0.991 and 0.994 and the values for MRE are 0.064 and 0.049, respectively. The audience should note that these visualizations are only a very small fraction of the total cases in the test dataset, thus, the metrics from the all test cases of the two models are more unbiased to compare models' performance.

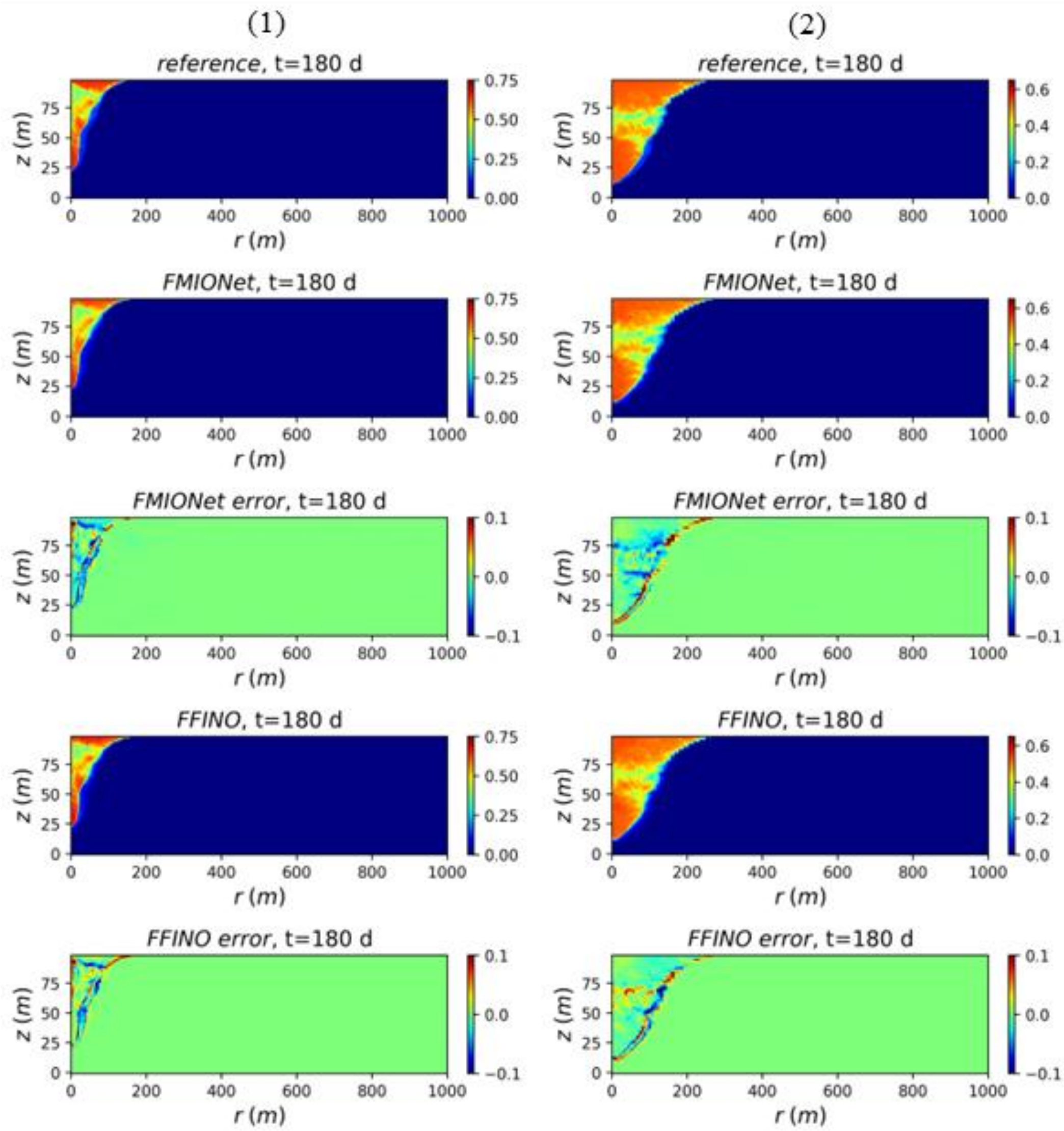

**Fig. 10**. Two test cases of FMIONet and FFINO for hydrogen saturation with each case presenting reference, predictions, and model errors on the 180$^{th}$ day

### *4.2. Pressure buildup*

Similar to the saturation study, FMIONet and FFINO are also employed to learn the mapping from input parameters to pressure buildup evolution. Table 5 compares the performance of the two models on pressure buildup predictions. The prediction accuracy on pressure buildup between FFINO and FMIONet is very close. Compared with FMIONet, FFINO is better and more stable across the selected metrics. FFINO achieves a 16.3% prediction accuracy improvement on RMSE and smaller standard deviations on all metrics. Both models exhibit much higher prediction accuracy for pressure buildup than for hydrogen saturation, which agrees with other studies on similar subsurface multiphase flow problems [16, 25].

**Table 5**. FFINO and FMIONet model performance on pressure buildup

| *Models* | *$R^2$* | *RMSE* | *SSIM* | *MRE* |
|---|---|---|---|---|
| *FMIONet* | 0.9988±0.0003 | 1.11e-1±1.11e-2 | 9.996e-1±4.12e-5 | 5.62e-2±4.21e-3 |
| *FFINO* | 0.9992±0.0001 | **9.29e-2±6.98e-3** | 9.997e-1±2.46e-5 | 5.47e-2±3.67e-3 |

Then, the accuracy of the best FMIONet and FFINO models on pressure buildup is presented. Fig. 11 gives the $R^2$ and MRE distributions of 200 cases from the pressure buildup for the two best trained models. It is obvious to conclude that our FFINO model outperforms FMIONet in $R^2$, while the two models are very close in terms of MRE, as shown in Fig. 11b. From the MRE distribution, the performance of the FFINO model is more stable across the test cases as the MRE values are more centered compared to those of FMIONet.

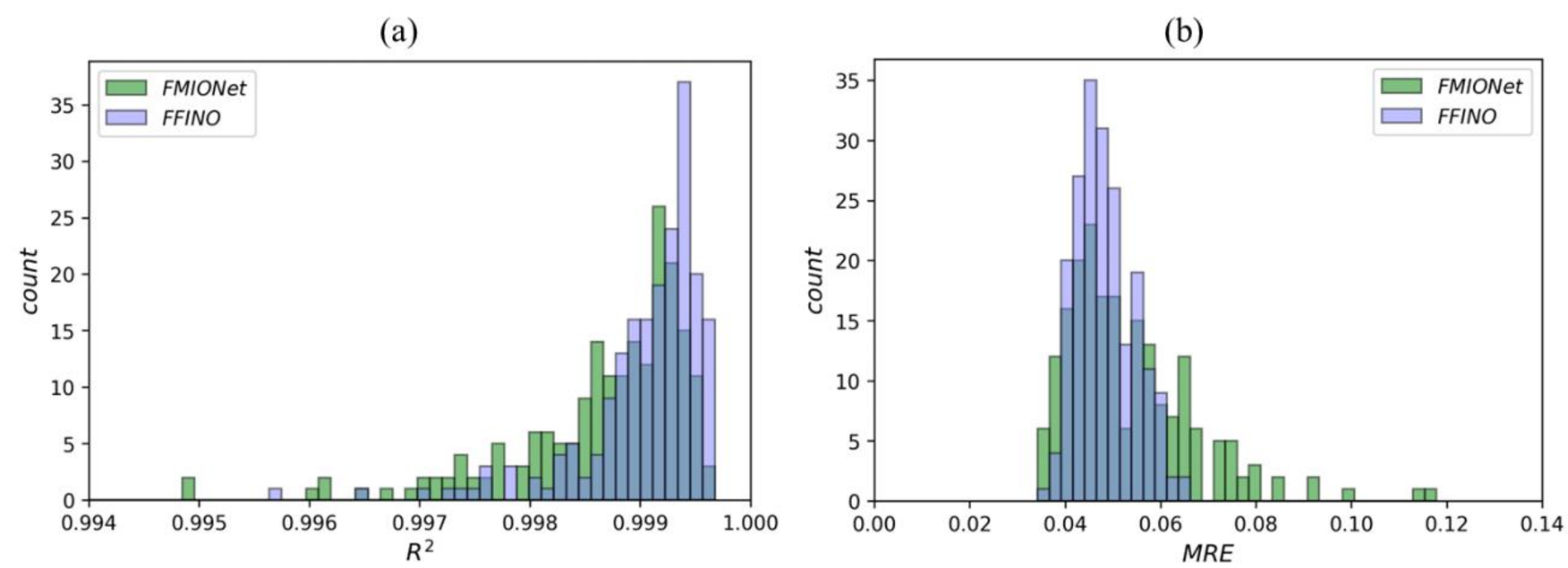

**Fig. 11**. $R^2$ and MRE distributions of 200 cases in pressure buildup testing for the best-trained FMIONet and FFINO models (a: $R^2$, b: MRE)

Visualizations of the 2 corresponding pressure buildup test cases, as in the hydrogen saturation, are presented in Fig. 12. Both FFINO and FMIONet models can predict pressure buildup with high accuracy. For the two cases, the $R^2$ scores of the two models are almost identical, both achieving values above 0.996. For MRE, FMIONet achieves 0.045 and 0.056 for the two cases respectively, while the values for FFINO are 0.058 and 0.060, which are still very close. As in the case of hydrogen saturation, the performance of the models should be evaluated on the whole test dataset rather than on limited visualized cases.

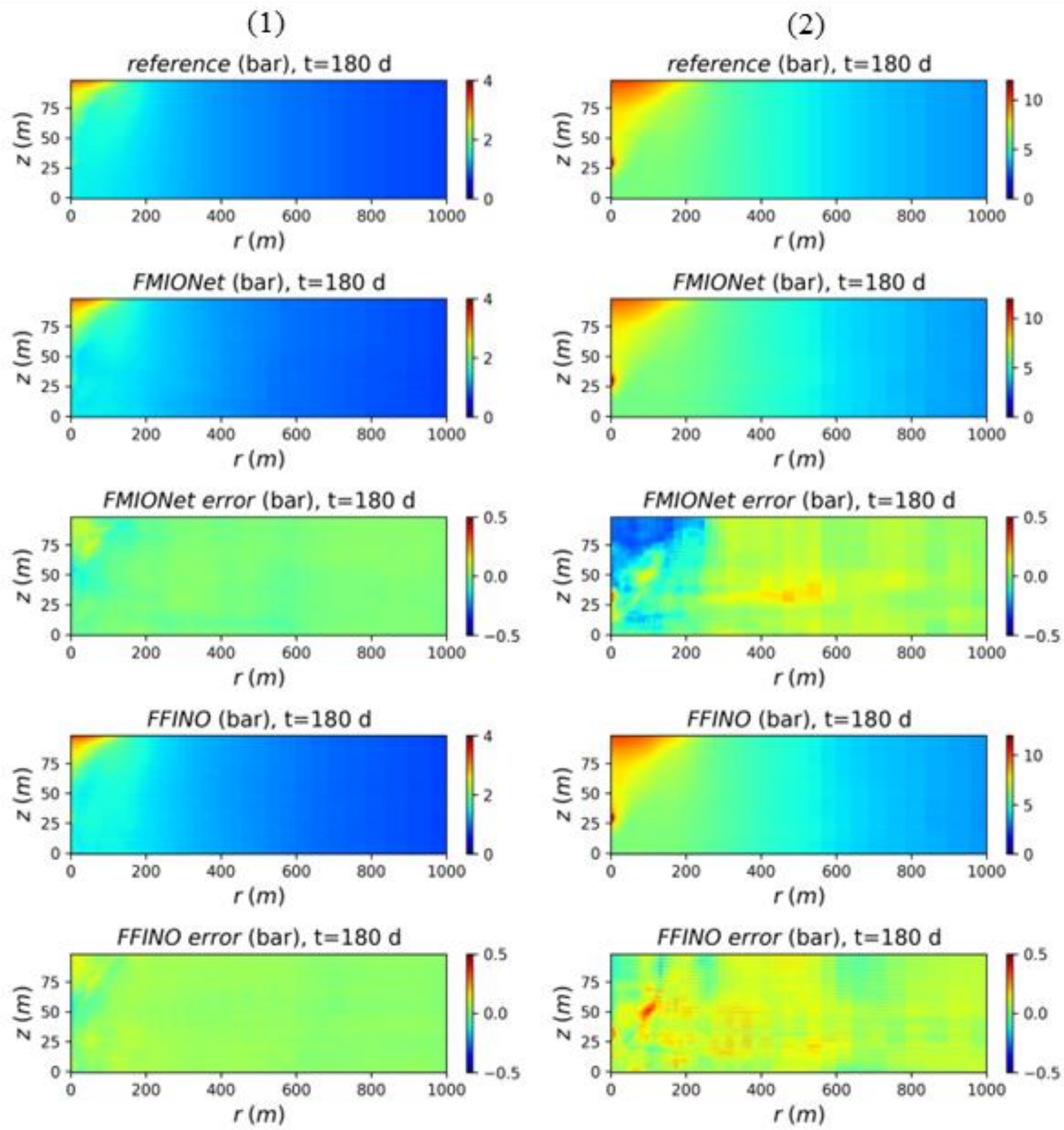

**Fig. 12**. Two test cases of FMIONet and FFINO for pressure buildup with each case presenting reference, predictions, and model errors on the 180$^{th}$ day

## *4.3. Computational efficiency analysis*

The comparison of FFINO and FMIONet model computational efficiency on training and testing is presented in Table 6. All experiments (both training and inference) are performed on a workstation with an NVIDIA RTX 5000 Ada Generation GPU with 32 GiB memory.

Compared with FMIONet, the most striking advantage is that FFINO reduces the number of trainable parameters from 3.69 million to 2.28 million, a 38.1% reduction.

Due to the smaller number of trainable parameters, a trained FFINO model requires 43.4% less storage compared to a trained FMIONet model.

**Table 6**. Comparison of FFINO and FMIONet model characteristics

| *Models* | *Number of Params.* | *Model size (MiB)* | *GPU memory (GiB)* | *Train time (s/epoch)* | *Inference time (s)* |
|---|---|---|---|---|---|
| *FMIONet* | 3,685,325 | 77.93 | 1.75±0.05 | 57.56±1.70 | 0.022±0.001 |
| *FFINO* | **2,282,981** | **44.07** | 1.54±0.05 | **47.43±0.62** | 0.020±0.001 |

During the training, FFINO requires 12% less GPU memory in comparison to FMIONet. The FFINO model is on average 17.6% faster than the FMIONet model on training time, namely 33.8 minutes faster for a 200-epoch training experiment. The inference time of the trained models is computed using the average prediction time of 200 test cases. The results indicate that the two models are comparable, and the FFINO model is 9% faster in inference. The inference time is approximately 7,850 times faster than the average runtime (157 seconds) of all 3,200 numerical simulations.

### *4.4. Sensitivity analysis*

As described in Section 2, the input parameters are categorized into spatial inputs ($k_h$, $k_v/k_h$, $\phi$) and scalar inputs ($Q$, $k_{rw,max}$, $k_{rg,max}$, $S_{wi}$, $S_{gr}$, $m$, $n$). To quantify how sensitive model performance is with respect to variations of input parameters, a One-Factor-At-a-Time (OAT) method is leveraged to assess model performance [52, 53]. The OAT method adds a proportional random Gaussian noise $\mathcal{N}(0,\ 0.25)$ to one input parameter while keeping all other input parameters fixed. The noise-free MREs for the best FMIONet and FFINO models on hydrogen saturation are 0.090 and 0.084, respectively, and the corresponding MRE values for pressure buildup are 0.054 and 0.048, respectively. The proposed OAT workflow iterates the process over the input parameter spaces to acquire the input parameters with noise and load the best-trained FMIONet and FFINO models to give new predictions. The process is repeated 5 times to eliminate random errors, and MRE values are computed to quantify model performance degradation after noise addition. The results of the OAT sensitivity analysis of input parameters are given in Table 7.

**Table 7**. Sensitivity analysis of input parameters on FMIONet and FFINO model performance via MRE

| ***Input Parameters*** | ***FMIONet*** | | ***FFINO*** | |
|---|---|---|---|---|
| | ***sg*** | ***dP*** | ***sg*** | ***dP*** |
| $\boldsymbol{k_h}$ | 0.093±0.0002 | 0.054±7.3e-5 | 0.087±9.7e-5 | 0.049±4.6e-5 |
| $\boldsymbol{k_v/k_h}$ | **0.098±0.0001** | 0.054±6.7e-5 | **0.096±0.0003** | 0.049±6.8e-5 |
| $\boldsymbol{\phi}$ | **0.098±0.0003** | 0.054±3.6e-5 | **0.094±0.0002** | 0.049±4.1e-5 |
| $\boldsymbol{Q}$ | **0.135±0.0029** | **0.170±0.0046** | **0.130±0.0027** | **0.159±0.0067** |
| $\boldsymbol{k_{rw,max}}$ | 0.094±0.0002 | **0.068±0.0016** | 0.088±0.0003 | **0.064±0.0023** |
| $\boldsymbol{k_{rg,max}}$ | **0.100±0.0008** | 0.056±0.0001 | **0.094±0.0013** | 0.051±0.0001 |
| $\boldsymbol{S_{wi}}$ | **0.102±0.0009** | 0.054±0.0001 | **0.097±0.0011** | 0.049±6.9e-5 |
| $\boldsymbol{S_{gr}}$ | 0.094±0.0004 | 0.054±4.8e-5 | 0.088±0.0002 | 0.049±6.7e-5 |
| $\boldsymbol{m}$ | **0.110±0.0016** | 0.057±0.0006 | **0.105±0.0010** | 0.052±0.0002 |
| $\boldsymbol{n}$ | **0.106±0.0011** | 0.057±0.0004 | **0.103±0.0012** | 0.052±0.0004 |

Both models exhibit very similar responses to input parameter variations. For hydrogen saturation and pressure buildup, the injection rate $Q$ is the most influential parameter to models' performance as it accounts for the largest MRE increase. For hydrogen saturation, models' performance is moderately sensitive to other parameters such as $m$, $n$, $S_{wi}$, $k_{rg,max}$, $k_v/k_h$, and $\phi$. However, for pressure buildup, $k_{rw,max}$ shows a moderate impact on models' performance, while other parameters are less influential.

# 5. Discussion

## *5.1. Paired t-tests*

Although results from Section 4 demonstrate FFINO's advantages over FMIONet from multiple aspects, further comprehensive paired t-tests are conducted to confirm the prediction accuracy advantage of FFINO over FMIONet.

A paired t-test can be used to determine whether statistically the difference of performance metrics for the two models is significantly different from zero. The procedure for the paired t-tests in our study is as follows:

1) The difference ($d$) of performance metrics (i.e., RMSE and MRE) for two models is computed on all test cases.
2) The mean ($\bar{d}$) and standard deviation ($s_d$) of the difference are calculated.
3) A null hypothesis ($H_0: \mu_d = 0$) is proposed, where $\mu_d$ is the population mean of the difference.
4) Compute $t$ scores and $p$ values with equations (13) and (14).
5) Compare $p$ values with a pre-selected significance level $\alpha = 0.05$.
6) Reject the null hypothesis if $p \leq \alpha$, and if $p > \alpha$, the null hypothesis not rejected.

$$t = \frac{\bar{d}}{s_d / \sqrt{n}}, \tag{13}$$

$$p = 2 \cdot \left[1 - F_{t,df}(|t|)\right], \tag{14}$$

where $n$ is the number of pairs, $df$ is the degree of freedom, which is $n-1$ for a paired t-test, and $F_{t,df}$ is the cumulative distribution function of the $t$ distribution with $df$.

Consequently, in our study, if the null hypothesis is rejected, which indicates that the metrics improvement of FFINO over FMIONet is statistically significant, otherwise, the metrics improvement is only some random variations. Two metrics RMSE and MRE on both hydrogen saturation and pressure buildup are selected. For hydrogen saturation, the average p-values of metrics RMSE and MRE are 4.72e-14 and 3.22e-14, respectively, which are significantly lower than the pre-selected significance level $\alpha$, thus rejecting the null hypothesis. Similarly, the average p-values of metrics RMSE and MRE on pressure buildup are 1.40e-6 and 0.023, respectively, which are also lower than the pre-selected significance level $\alpha$, thus rejecting the null hypothesis. The paired t-tests indicate that we are at least 95% confident that the prediction accuracy improvement of FFINO over FMIONet is statistically significant.

### 5.2. Limitations and future work

The current work is based on a synthetic 2D aquifer model, which is an idealized case compared to the real-field scenarios. In real-field applications, $H_2$ is typically injected from the top of the anticline (or seal) to minimize hydrogen loss [4, 6, 39, 40, 44]. In addition, we only consider the injection stage and drainage relative permeability curves for this study, which are not representative enough for UHS applications.

Although the FFINO model is developed based on the context of UHS, it is a general model applicable to other subsurface applications, such as GCS and oil and gas extraction. To make the FFINO model more suitable for UHS applications, cyclic injection and withdrawal schemes, and relative permeability hysteresis should be included in future work.

## 6. Conclusions

In this work, a new deep neural operator architecture, FFINO, is proposed to model hydrogen injection in saline aquifers. The main conclusions of the study is summarized below.

The simulation study with various relative permeability curves reveals that even a nuance of relative permeability can significantly impact the results of simulations. It is confirmed that the buoyancy effect of the hydrogen plume is strong, and the horizontal extent of the plume is limited until it reaches the top boundary. The finding from the simulation study pinpoints the importance of accurate relative permeability information for UHS simulations.

The comparisons show that the FFINO model is 9.8% more accurate for hydrogen plume prediction on MRE and 16.3% more accurate for pressure buildup prediction on RMSE than the FMIONet model which is also supported by paired t-tests. The FFINO model also exhibits more stable performance across all the testing metrics. Compared to FMIONet architecture, FFINO architecture has 38.1% fewer training parameters and consumes 12% less GPU memory. The training of FFINO is 17.6% faster, and the inference of FFINO is 9% faster.

Sensitivity analysis shows that the most influential parameter on models' performance is injection rate $Q$. Parameters like $m$, $n$, $S_{wi}$, $k_{rg,max}$, $k_v/k_h$, and $\phi$ account for moderate impact on models' performance for hydrogen saturation. While for pressure buildup, $k_{rw,max}$ shows a moderate impact.

The superior performance of FFINO facilitates fast and accurate estimates of plume migration and pressure distribution, reducing the time cost of decision making for UHS applications, and it can also be integrated into history matching and optimization pipelines to replace computationally expensive numerical solvers and deliver excellent time efficiency.

## Code and data availability

The code and datasets used in this paper will be available on GitHub at https://github.com/TaoWang417/ffino-uhs upon publication of the paper.

## Authorship contribution statement

**Tao Wang:** Writing – review & editing, Writing – original manuscript, Visualization, Validation, Software, Simulation, Code, Methodology, Investigation, Formal analysis, Data curation. **Hewei Tang:** Conceptualization, Funding Acquisition, Writing – review & editing, Writing – original manuscript, Supervision, Resources, Methodology.

## Acknowledgments

The work is supported by the startup funding from the Cockrell School of Engineering at the University of Texas at Austin (UT Austin). We also thank Dr. Mojdeh Delshad and her PhD student, Hussain Sadam, for their generous help with the CMG simulation setup. The authors acknowledge Computer Modeling Group Ltd. for providing CMG package licenses.

Appendix A. Statistics of the sampled spatial and scalar parameters in the study as shown in Table A.1

**Table A.1**

| Type | Variable | Max | Min | Mean | Std | Unit |
|---|---|---|---|---|---|---|
| spatial | $k_h$ | 1,000.0 | 44.1 | 327.0 | 132.8 | mD |
| | $k_v/k_h$ | 1.00 | 0.01 | 0.305 | 0.134 | – |
| | $\phi$ | 0.345 | 0.140 | 0.242 | 0.024 | – |
| scalar | $Q$ | 255,000 | 25,500 | 135,000 | 65,450 | $m^3$/day |
| | $k_{rw,max}$ | 0.768 | 0.530 | 0.649 | 0.069 | – |
| | $k_{rg,max}$ | 0.056 | 0.031 | 0.043 | 0.007 | – |
| | $S_{wi}$ | 0.500 | 0.340 | 0.421 | 0.046 | – |
| | $S_{gr}$ | 0.120 | 0.030 | 0.075 | 0.026 | – |
| | $m$ | 3.808 | 1.454 | 2.628 | 0.676 | – |
| | $n$ | 3.317 | 1.052 | 2.185 | 0.658 | – |